\documentclass{article}

\PassOptionsToPackage{numbers,compress}{natbib}
\usepackage[dblblindworkshop, final]{neurips_2026}
\workshoptitle{Foundation Models for the Brain and Body}

\usepackage[utf8]{inputenc}
\usepackage[T1]{fontenc}
\usepackage{hyperref}
\usepackage{url}
\usepackage{booktabs}
\usepackage{amsfonts}
\usepackage{amsmath}
\usepackage{nicefrac}
\usepackage{microtype}
\usepackage{xcolor}
\usepackage{multirow}
\usepackage{tikz}
\usetikzlibrary{arrows.meta,positioning,fit}
\usepackage{graphicx}
\graphicspath{{figures/}}

\newcommand{\nz}{ZarrFM}
\newcommand{\code}[1]{\texttt{#1}}

\title{ZarrFM: A Canonical Tensor Storage Architecture for EEG Foundation-Model Pretraining}

\author{
  \textbf{Dheeraj Chintapalli}$^{1,3\thanks{Equal Contribution}}$ \quad
  \textbf{Sunkalp Chandra}$^{2,3^{*}}$ \\[1ex]
  $^1$ University of Michigan, Ann Arbor \\
  $^2$ University of Texas, Austin \\
  $^3$ Reteena \\
  \texttt{dheerajian@umich.edu}, \texttt{sunkalp@utexas.edu}
}

\begin{document}

\maketitle

\begin{abstract}
Statistical models for brain-computer interfaces (BCIs) increasingly require large, heterogeneous EEG corpora, yet existing datasets vary in sampling rates, channel layouts, window lengths, and storage formats, creating substantial preprocessing and I/O overhead. We propose \textsc{ZarrFM}, an execution-oriented representation that complements BIDS by storing a corpus as a single Zarr v3 hierarchy with indexed metadata, canonical channel spaces, explicit masks, hash-stable subject splits, verified event tables, and block-aligned sharding. On real PhysioNet recordings and synthetic corpora, we compare \textsc{ZarrFM} with EDF, FIF, NPY, HDF5, and Parquet. Across 200 recordings, its two-level index reduces metadata queries from 601 requests to 14. Preserving native signal codes reduces storage from 42.4 MB/hour in EDF to 22.7 MB/hour, while block-aligned sampling reaches 3,131 windows/s, up to $11\times$ faster than Parquet. Reads remain bit-identical across independent Zarr implementations, and concurrent failure-injected appends recover without data loss. These results show that \textsc{ZarrFM} can provide a compact, interoperable, and execution-efficient storage layer for large-scale EEG pretraining.

\end{abstract}

\section{Introduction}
Recent EEG foundation models, including BENDR~\citep{kostas2021bendr}, BIOT~\citep{yang2023biot}, LaBraM~\citep{jiang2024labram}, EEGPT~\citep{wang2024eegpt}, CBraMod~\citep{wang2025cbramod}, REVE~\citep{elouahidi2025reve}, pretrain on increasingly large corpora, ranging from 2,500 to 61,000 hours~\citep{jiang2024labram,jiang2025neurolm,elouahidi2025reve}. However, CBraMod's gains slow beyond 1,000 hours, and recent surveys find limited evidence that current models benefit reliably from further scaling~\citep{wang2025cbramod,kuruppu2025eegfmreview,liu2026eegfmcompass}. Yet these corpora are not directly ingestable by models, as each training pipeline independently parses EDF~\citep{kemp1992edf,kemp2003edfplus} or BIDS~\citep{gorgolewski2016bids,pernet2019eegbids}, resolves channel identities, resamples and re-references signals, segments windows, handles missing electrodes, and assigns subject-level splits. Implementations then materialize their own training representations: CBraMod retains 34\% of TUEG in an LMDB of pickles, NeuroLM writes one pickle per window, and EEG-FM-Bench stores pre-cut windows in Parquet~\citep{wang2025cbramod,jiang2025neurolm,xiong2026eegfmbench}. Even when using the same recordings, preprocessing differs: three models band-pass TUEG at 0.1--75, 0.3--75, and 0.5--50 Hz~\citep{jiang2024labram,wang2025cbramod,chen2026unintfm}. These choices alter the numerical training distribution and make comparisons difficult to reproduce. Recent surveys consequently identify data standardisation as an open problem, including the need for consistent subject-level splits~\citep{liu2026eegfmcompass,kuruppu2025eegfmreview}.

A training corpus must therefore support three operations. \textbf{Q1}~\emph{Harmonize:} channel identity, sampling rate, and subject splits should be resolved once and shared across models. \textbf{Q2}~\emph{Read:} long, multichannel recordings should support efficient window access from local or remote storage without starving accelerators. \textbf{Q3}~\emph{Maintain:} annotations and channel-level quality information should be added without rewriting signal data or creating inconsistent state. BIDS provides an archival and interoperability layer, but is not designed around these execution requirements. We therefore propose an \emph{execution representation} alongside the archival representation: a corpus is converted once into a training-ready form rather than repeatedly translated by individual pipelines, following the broader approach of OME-Zarr~\citep{moore2023omezarr}. Canonical channel identities, tensorized metadata, chunked signals, subject-level splits, and sharded sampling follow directly from these requirements.

\paragraph{Contributions.} (i)~A normative specification of \nz\ (\S\ref{sec:design}, Appendix~\ref{app:schema}) using only core Zarr~v3 types and codecs, with explicit conformance rules. (ii)~A two-level consolidated index that makes corpus initialization $O(\text{index})$ and recording access a single request. (iii)~A block-aligned sharded sampler with a PyTorch adapter providing exactly-once window coverage across distributed ranks and workers. (iv)~An append-only edit protocol for closed stores that avoids in-place modification, validated through crash injection and concurrent-writer tests. (v)~A systematic evaluation against EDF, FIF, NPY, HDF5, and Parquet on real PhysioNet recordings and synthetic corpora, including a corpus-scale conversion of all open-access PhysioNet EDF databases (236.5 GB, 11,877 recordings) that is bit-exact across two independent Zarr implementations.

\section{Background}

\emph{Standards and containers.} EDF defines the int16 signal representation and physical-unit calibration we preserve~\citep{kemp1992edf,kemp2003edfplus}; BIDS-EEG organizes recordings via JSON/TSV sidecars but does not define signal encoding~\citep{pernet2019eegbids,appelhoff2019mnebids}. NWB uses an HDF5-based, session-oriented container~\citep{rubel2022nwb,tritt2019hdmf}, while IBL's ONE provides array revision and access mechanisms with a separate raw-signal representation~\citep{ibl2023one}. These systems support archival organization and interchange, but not corpus-level representation or training-time indexing.

\emph{Cloud-native arrays.} OME-Zarr~\citep{moore2023omezarr} and VCF-Zarr~\citep{czech2025vcfzarr} demonstrate scientific data representation with chunked Zarr arrays~\citep{zarr2023v3spec,abernathey2021cloudnative}. Electrophysiology studies have examined Zarr compression~\citep{buccino2023compression} and object-storage latency and cost~\citep{durner2023cloudstorage}, but leave corpus indexing and sampling to downstream systems.

\emph{Loaders and EEG pipelines.} WebDataset~\citep{aizman2019webdataset}, FFCV~\citep{leclerc2023ffcv}, and CorgiPile~\citep{xu2022corgipile,livne2023corgi2} optimize sample delivery through sharding and block-level access, while prior work identifies input preparation as a training bottleneck~\citep{mohan2021datastalls,isenko2022bottleneck}. EEG pipelines additionally face montage and sampling-rate heterogeneity~\citep{liu2026eegfmcompass}, incomplete dataset standardization~\citep{aristimunha2026eegdash}, and leakage from segment-level evaluation~\citep{brookshire2024leakage,delpup2025partitioning}. \nz\ targets the layer between archival formats and model code: standard Zarr tooling remains usable, while its index, layout, and edit protocol provide a common representation for corpus-scale sampling and maintenance. MNE~\citep{gramfort2013mne} provides the conversion bridge.

\section{Design}\label{sec:design}

\begin{figure}[t]
\centering
\begin{minipage}[t]{0.50\textwidth}
\vspace{0pt}
\scriptsize
\begin{tabular}{@{}l@{}}
\code{corpus.zarr/}\\
\code{~~recordings/sub-001\_ses-01/}~~\textit{one group per recording}\\
\code{~~~~signals/eeg}~~~~$(C,N)$ int16, chunks $(C,{\sim}1\,\text{MiB})$, sharded\\
\code{~~~~channels/eeg/}~~name, type, unit, reference (JSON attrs);\\
\code{~~~~~~~~~~~~~~~~~~}gain, offset, bad, \textbf{canonical\_id}, position\\
\code{~~~~events/r$k$/}~~~~onset, duration, value, \textbf{label\_id}; digest, vocab\\
\code{~~~~provenance/}~~~~source, pipeline, software\\
\code{~~index/}~~~~~~~~~~~~$(R,)$ n\_samples, sfreq, split, content\_hash, \ldots\\
\code{~~~~canonical\_present}~$(R,K)$ bool, good slots\\
\code{~~subjects/ splits/ vocab/ canonical/ provenance/}\\
\end{tabular}
\end{minipage}\hfill
\begin{minipage}[t]{0.47\textwidth}
\vspace{0pt}
\centering
\begin{tikzpicture}[node distance=1.6mm, every node/.style={font=\scriptsize}, box/.style={draw, rounded corners=1pt, inner sep=1.6pt, minimum width=3.6cm, align=center}, arr/.style={-{Latex[length=1.2mm]}, thick}]
\node[box] (store) {object store / NFS / NVMe\\ shard ($\approx$64\,MiB) $\supset$ chunk ($\approx$1\ MiB, all ch.)};
\node[box, below=of store] (reader) {\code{DatasetReader}: 1 request $\to$ index, splits, vocab};
\node[box, below=of reader] (sampler) {\code{WindowSampler}: window table $\to$ blocks\\ shuffled per epoch, dealt to (rank, worker)};
\node[box, below=of sampler] (read) {read block once $\to$ project $(C,T)\!\to\!(K,T)$ + mask\\ slice windows; events window-relative};
\node[box, below=of read] (gpu) {collate $(B,K,T)$, pinned H2D, $X\odot\text{mask}$ $\to$ model};
\draw[arr] (store) -- (reader); \draw[arr] (reader) -- (sampler); \draw[arr] (sampler) -- (read); \draw[arr] (read) -- (gpu);
\end{tikzpicture}
\end{minipage}
\caption{\textbf{Left:} The \nz\ hierarchy: one corpus and one store containing $R$ recordings and $K{=}112$ canonical channel slots, with \code{r$k$} denoting the current event revision. \textbf{Right:} The training path, with subjects and sessions stored as index columns rather than groups.}
\label{fig:overview}
\end{figure}
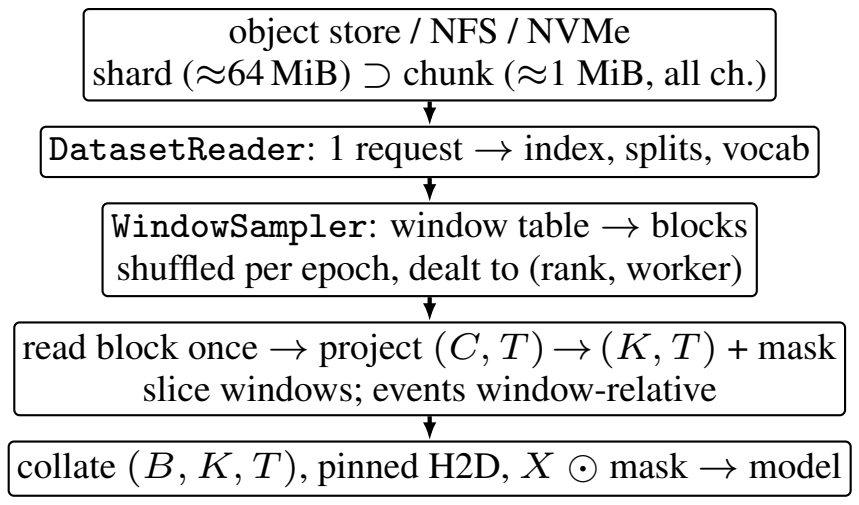

\paragraph{Canonicalization.} A corpus is represented as a single Zarr~v3 store (Fig.~\ref{fig:overview}); each recording is a self-describing group containing channels-major int16 signals, with one array per native sampling rate. Channels remain in \emph{physical acquisition space}, retaining vendor name, type, unit, reference, position, and bad-channel status, and receive a \texttt{canonical\_id} into a frozen, versioned \emph{canonical model space}: 91 extended 10--10 scalp positions~\citep{oostenveld2001fivepercent,jurcak2007tenten} and 21 auxiliary classes (Appendix~\ref{app:canonical}). Canonicalization is a deterministic function of channel names, types, and table version, evaluated once at ingestion: \texttt{EEG FP1-REF}\,$\to$\,\texttt{Fp1} with reference \texttt{REF}, \texttt{T3}\,$\to$\,\texttt{T7}, and \texttt{C3-M2}\,$\to$\,\texttt{C3} with reference \texttt{M2}; intracranial channels map to $-1$. The mapping is an index, not interpolation: loading scatters physical channels from $(C,T)$ into canonical space $(K,T)$ with a $(K,)$ presence mask, while retaining the original channel metadata for set- and coordinate-based models~\citep{yuan2024brainwave,elouahidi2025reve}. When multiple physical channels map to one canonical slot, the collision is resolved deterministically and the retained \texttt{reference} records the discarded context. Event onsets are float64 and labels use a dataset-wide vocabulary constructed at ingestion, so each \texttt{label\_id} has corpus-wide semantics. Subject splits are deterministic hashes of $(\text{seed},\text{subject})$ against cumulative split fractions; adding recordings therefore cannot move an existing subject, and validation rejects manifests containing cross-split subject leakage. The loader performs no implicit resampling, re-referencing, or channel inference: signals remain separated by native sampling rate, and incompatible mixed-rate selections fail validation. These invariants are materialized in \texttt{index/}: one aligned row per recording containing length, native rate, subject, split, byte count, and SHA-256 content hash, together with a $(R,K)$ \texttt{canonical\_present} matrix. Corpus queries can therefore resolve recordings and canonical coverage directly from the index without opening signal objects.

\paragraph{Sampling.} Signal arrays are chunked across the full channel dimension and approximately $1$~MiB of time, with chunk sizes constrained to $2$--$64$~s, $256$~KiB--$4$~MiB, and at most one eighth of the recording (\S\ref{sec:exp}); chunks are grouped into approximately $64$~MiB shards. Zarr consolidated metadata normally places node documents in a root JSON object, duplicating approximately $17R$ recording-level documents and forcing readers to parse unrelated metadata. \nz\ instead uses two-level consolidation: each recording retains its complete local document, while the root contains only dataset metadata and the corpus index. Opening a corpus therefore requires one $O(\text{index})$ request, with one additional metadata request per recording. 

\texttt{WindowSampler} represents windows as integer tuples and groups them by backing storage block rather than shuffling windows independently. Blocks correspond to complete chunks, with adjacent chunks optionally coalesced into approximately $8\,\text{MiB}$ reads. Each epoch deterministically permutes blocks from $(\text{seed},\text{epoch})$ and assigns them round-robin across $\text{world\_size} \times \text{num\_workers}$ consumers. Each worker reads and projects each assigned block once, then emits its windows in shuffled order. This 
shares decompression and object-store reads across windows while providing exactly-once coverage without shared sampler state. The resulting block-level 
shuffle is analogous to CorgiPile~\citep{xu2022corgipile,livne2023corgi2}, with 
the storage chunk as the shuffle unit. 

\paragraph{Consistency.} Post-ingestion annotations are represented as immutable, SHA-256-digested event revisions~\citep{ibl2023one}, where a writer claims $\texttt{events/r}_{k+1}$ with set-if-not-exists, which is written and verified under used keys, and committed through a single atomic root-document update~\citep{zarr2023v3spec,durner2023cloudstorage}. This leads to readers seeing either the old or new full table without a partial one, and channel quality flags and provenance update attributes without rewriting signals~\citep{ibl2023one}.

\section{Experiments}\label{sec:exp}

\paragraph{Setup.} We evaluate \nz{} on a $64$-core Linux host with local NVMe storage, Python~$3.12$ (uv), zarr-python~$3.3.0$, PyTorch~$2.8$ (cu128), and seven Quadro RTX~$6000$ GPUs. We use three representative PhysioNet recordings derived from datasets used by prior EEG foundation models and benchmarks~\citep{jiang2024labram,yuan2024brainwave,kastrati2025eegbench}: CHB-MIT \texttt{chb01\_03} ($23$ bipolar channels, $256\,\text{Hz}$, $1\,\text{h}$, $12$-bit EDF), EEGBCI \texttt{S001R04} ($64$ channels, $160\,\text{Hz}$, $2\,\text{min}$, EDF+ annotations), and Sleep-EDF \texttt{SC4001E0} (three $100\,\text{Hz}$ EEG/EOG channels, four $1\,\text{Hz}$ auxiliary channels, $22\,\text{h}$, expert hypnogram). Synthetic EEG with volume conduction, alpha activity, blinks, ERP, and anti-alias filtering is used for experiments requiring controlled variation across recordings. For corpus-level compression, we include every open-access PhysioNet database distributed in EDF: $15$ databases comprising $12{,}377$ files ($11{,}877$ signal recordings and $500$ annotation-only scorer files), $8{,}359$ hours and $236.5$ GB. Data are retrieved from the \texttt{physionet-open} S3 mirror and converted using the byte-shuffle \texttt{fast} profile with warm runs. Because the corpus contains integer-coded signals, we additionally use one float64 MOABB subject (BNCI2014\_009, P300) to evaluate non-integer input handling.


\begin{table}[t]
\caption{Random-window reads on one CHB-MIT clinical hour (23 ch, 256 Hz): 512 random 4 s windows, all channels, one process, local NVMe; medians of four warm runs. NPY is an uncompressed memmap; Parquet stores pre-cut float32 windows with zstd, following EEG-FM-Bench~\citep{xiong2026eegfmbench}, with 16-window row groups.}
\label{tab:formats}
\centering\footnotesize
\setlength{\tabcolsep}{5pt}
\begin{tabular}{@{}lrrrr@{}}
\toprule
format & size & windows/s & MB/s out & ms/window \\
\midrule
NPY float32 memmap (ceiling) & 84.8\,MB & \textbf{50031} & 4713 & 0.02 \\
HDF5 float32, chunked, uncompressed & 84.8\,MB & 15786 & 1487 & 0.06 \\
\textbf{\nz, block-merged reads} & \textbf{22.7\,MB} & \textbf{3131} & 295 & 0.32 \\
HDF5 float32, gzip-4 & 33.9\,MB & 1584 (1055--1615) & 149 & 0.63 \\
FIF (MNE, \code{preload=False}) & 84.8\,MB & 672 & 63 & 1.49 \\
EDF, original file (MNE, \code{preload=False}) & 42.4\,MB & 595 & 56 & 1.68 \\
Parquet, materialised 4\,s windows, zstd & 50.0\,MB & 279 & 26 & 3.6 \\
\nz, one read per window & 22.7\,MB & 168 (167--378) & 16 & 2.6--6.0 \\
\bottomrule
\end{tabular}
\end{table}

\paragraph{I/O throughput (Table~\ref{tab:formats}).} With block-merged reads, \nz\ is $5\times$ faster than EDF, $4.7\times$ than FIF, $2\times$ than compressed HDF5, and $11\times$ than Parquet storing pre-cut windows, while using half the EDF storage and one quarter of float32. Uncompressed formats are another $5$--$16\times$ faster, reflecting the cost of decompression. With one request per window, \nz\ is instead the slowest format at 2.6--6 ms per window, dominated by zarr-python's fixed request overhead rather than I/O. A 4 s window is only 47 KB, making per-sample access disproportionately expensive~\citep{isenko2022bottleneck}. The block sampler is therefore essential: merging reads improves throughput by $8$--$19\times$; without it, \nz\ incurs the per-window overhead above.

\paragraph{Metadata \& Requests (Table~\ref{tab:metadata}, Appendix~\ref{app:more}).} On local NVMe, parsing 601 JSON/TSV files takes 140 ms while tensorized metadata without aggregation requires 13 requests per recording and runs $48\times$ slower than sidecars locally. Leveraging consolidated index leads to metadata access at 14 requests, instead of $3R{+}1$ for sidecars and $\sim13R$ for unaggregated tensor metadata, yielding $4\times$ lower latency on NVMe and $43\times$ at 10 ms/request. Signal access shows the same effect: on the clinical hour, block sampling reads 15 requests for 900 windows, while per-window access issues 900; at 30 ms/request, their overheads reduce throughput by 34\% and 80\%, respectively. The 10 and 30 ms regimes bracket measured S3 round trips and first-byte latency~\citep{ofeidis2024dataloader,durner2023cloudstorage}.

\begin{figure}[t]
\centering
\includegraphics[width=0.75\textwidth]{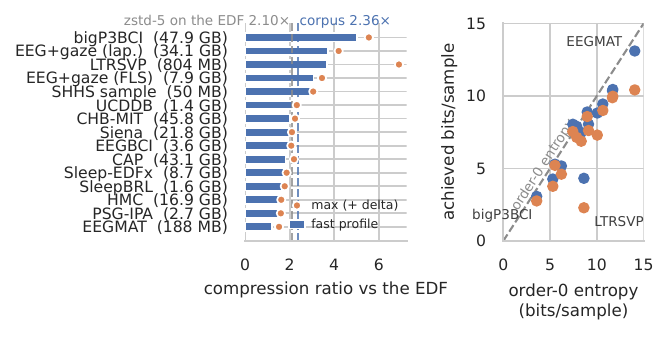}
\caption{All open-access PhysioNet EDF databases: 15 databases, 11,877 recordings, 236.5 GB, 8,359 h, with bit-exact readback. \textbf{Left:} byte-weighted compression relative to EDF; bars show \code{fast}, dots \code{max}, dashed lines corpus totals. \textbf{Right:} achieved bits/sample versus order-0 entropy. LTRSVP reaches first-difference entropy, while bigP3BCI's $5\times$ ratio reflects 82 boolean flag channels.}
\label{fig:corpus}
\end{figure}

\paragraph{Corpus-Scale Compression (Fig.~\ref{fig:corpus}).}
Recorders commonly quantize signal data to 12-bit precision within wider integer containers; however, \nz{} preserves the integer lattice without re-quantizing to full \texttt{int16}~\citep{buccino2023compression}. Across the 236.5\,GB PhysioNet corpus, \nz\ reduces storage to 100\,GB with the \texttt{fast} profile ($2.36\times$) and 89.8\,GB with \texttt{max} ($2.63\times$). While \texttt{max} saves an additional 10\% space, it reduces read throughput by one-third. In comparison, applying \texttt{zstd-5} directly to raw EDF files yields 112.6\,GB ($2.10\times$). All 11,877 recordings and 233,709 events are verified bit-exact upon readback.

Compression gains correlate strongly with bit depth as 12-bit recordings (comprising 80\% of dataset channels) yield higher compression ratios, whereas full 16-bit databases compress only $1.21$--$1.69\times$. At native sampling rates, \nz\ achieves $0.81$--$1.08\times$ the order-0 entropy of stored codes. Predictive coding could yield higher compression ratios~\citep{capurro2016efficientsequentialcompressionmultichannel}, but would compromise random access and portability. Stream-level shuffle auto-selection reduces corpus size by an additional 2.7\% (7.6\% on CHB-MIT) while improving write speeds by $\sim70\%$. For float sources lacking an integer lattice, \nz\ quantizes inputs to a user-defined step size; on a \texttt{float64} P300 subject discretized at $0.01\,\mu\text{V}$, this reduces footprint from 3.61\,MB (\texttt{float32}) to 1.03\,MB while tracking quantization parameters in the store provenance.

\paragraph{Conversion \& Analysis (Table~\ref{tab:real}).} 
\nz\ enables rapid, fault-tolerant ingestion, converting a 22-hour Sleep-EDF recording in $0.6\,\text{s}$. Although real-world metadata often breaks strict parsers, \nz\ safely ingests 227 CHB-MIT files with zero-span bounds and a CAP file with inverted ranges. Unlike loaders that implicitly resample mixed-rate data, \nz\ preserves native rates as distinct streams. For downstream analysis, \nz\ opens recordings $1.4$--$5\times$ faster than EDF header parsing and extracts grid-aligned epochs $3$--$7\times$ faster. However, because \nz\ optimizes for multichannel blocks, raw EDF remains faster for sparse, single-channel clinical viewing.


\paragraph{Portability.} We evaluated \nz\ against TensorStore (C++) and zarrs (Rust). Six string arrays from the earlier layout failed at parse time; all numeric nodes were bit-identical. Moving strings to attributes (\S\ref{sec:design}) allowed both implementations to read every array in every real store. A 93-line standalone reader using only the specified codecs reproduced all arrays directly from the specification. For edit safety, we injected failures after each of the 28--34 store operations in both new and migrating stores. Every fresh reader saw either the complete old table or complete new table; torn states were rejected by the validator, and \code{repair()} restored conformance without changing event counts. Concurrent appends from four processes (eight in review runs), with and without locking and alongside channel edits and readers, lost no acknowledged appends. Conflicting writers returned only the defined conflict error, while readers encountering invalid state returned only the integrity error. Three adversarial review rounds identified four additional write-loss cases (Appendix~\ref{app:protocol}).

\section{Limitations \& Conclusion}

We present ZarrFM, an execution-oriented tensor storage architecture designed to improve storage efficiency, access throughput, and data integrity for large-scale EEG foundation-model pretraining. The current implementation, however, is evaluated within specific system and representation boundaries. Although block sampling reduces request and decompression overhead, we do not evaluate how buffer size affects optimization dynamics, convergence, or statistical efficiency~\citep{xu2022corgipile,livne2023corgi2}. Second, the canonical projection is lossy for montages whose semantics depend on the reference or electrode geometry: omitting \code{reference} can substantially degrade performance on CHB-MIT~\citep{kastrati2025eegbench}, while name-based channel mapping does not preserve the spatial information required for geodesic nets, for which electrode positions must be retained~\citep{truong2023montages}. Third, float-valued sources without an integer lattice are quantized to a declared step size; although the step and quantization bound are recorded in provenance, this representation does not preserve the original values bit-exactly. Fourth, no end-to-end model was evaluated under \textsc{ZarrFM}: reported gains are storage size, metadata-query cost, and random-window loading throughput. Fifth, the throughput comparison confounds sampler and format, because only \textsc{ZarrFM} used block-aligned reads while baselines read one window at a time. Sixth, on bipolar clinical montages such as CHB-MIT, distinct channels (e.g., FP1-F7 and FP1-F3) can collide onto one canonical slot (Fp1), so one signal is dropped. 

These limitations define the scope of the current representation rather than failures of the storage abstraction itself. Future work should therefore evaluate end-to-end training under matched loaders, isolate sampler from format with block-aligned baselines, extend montage representations to preserve reference and spatial geometry, and quantify the effect of quantization and representation choices on downstream model performance.

\section{Schema Information}
Appendix~\ref{app:schema} provides a supplemental summary on the ZarrFM schema.

\newpage

\bibliographystyle{plainnat}
\bibliography{references}

\newpage
\appendix

\section{Schema}\label{app:schema}

\begin{table}[h]
\caption{Dataset-level requirements. Recording-level requirements are defined in the container specification released with the code.}
\centering\footnotesize
\begin{tabular}{@{}p{3.1cm}p{9.7cm}@{}}
\toprule
node & requirement \\
\midrule
root attrs & \code{format="zarrfm-dataset"}, \code{version} (semver), \code{dataset\_id} (UUID by default), \code{canonical\_version}, \code{n\_recordings}; SHOULD: \code{name}, \code{description}, \code{created}, \code{created\_by}. \\
\code{recordings/}\allowbreak\code{<id>/} & one conforming recording per id; id is one path component, SHOULD read \code{sub-X\_ses-Y[\_task-Z]}. \\
\code{index/} & \code{recording\_id}, \code{session}, \code{task}, \code{primary\_stream}, \code{content\_hash} as JSON list attrs; arrays \code{subject\_idx} int32, \code{n\_channels} int32, \code{n\_samples} int64, \code{sfreq}/\code{duration\_s} float64, \code{n\_events}/\code{nbytes} int64, \code{split} int8 ($-1$ unassigned), all length $R$; \code{canonical\_present} $(R,K)$ bool. MUST agree with the recordings' primary streams. \\
\code{subjects/} & attr \code{id} $(S,)$ unique; optional aligned float64 arrays or string-list attrs (minimal by policy). \\
\code{splits/} & attr \code{names}; attrs \code{level} $\in$ \{subject, session, recording\}, \code{seed}, \code{fractions}, \code{explicit}. No unit at the declared level may straddle splits. \\
\code{vocab/} & attr \code{event\_labels}; every recording's live revision \code{vocab} MUST equal it. \\
\code{canonical/} & attrs \code{name}, \code{type} $(K,)$; array \code{position} $(K,3)$ float64; attr \code{version}. \\
\code{provenance/} & attrs: \code{sources[]} (dataset, version, license, acquisition), \code{pipeline[]} (ordered steps with parameters), \code{software}, \code{written}. \\
consolidated metadata & at the root covering all nodes \emph{except} \code{recordings/*}; at each recording covering itself and only the live event revision. \\
\bottomrule
\end{tabular}
\end{table}

\begin{table}[h]
\caption{Recording-level channel and event tables (format 0.3). The default compression profile uses only Zarr~v3 core data types and codecs; the optional delta-filter profile is not portable. Missing values use NaN for position, $-1$ for the canonical id, \code{""} for reference, and \code{bad=true} for bad channels (data retained). Gaps are stored in \code{segments/}; samples are never NaN.}
\centering\footnotesize
\begin{tabular}{@{}p{4.1cm}lp{6.0cm}@{}}
\toprule
node & type & meaning \\
\midrule
\code{channels/*} attrs \code{name}, \code{type}, \code{unit}, \code{reference} & JSON lists $(C)$ & vendor name; type vocabulary; unit after scaling; what the channel was measured against \\
\code{channels/*/gain, offset} & float64 $(C,)$ & physical $=$ raw$\cdot$gain $+$ offset; the source's own when its codes fit int16 \\
\code{channels/*/bad} & bool $(C,)$ & flagged bad; masked, not removed; the one node an edit rewrites in place \\
\code{channels/*/canonical\_id} & int32 $(C,)$ & index into the canonical table, $-1$ if none \\
\code{channels/*/position} & float64 $(C,3)$ & head-frame metres (x right, y anterior, z up) \\
\code{events/r$k$/onset, duration} & float64 $(E,)$ & seconds from \code{t0}; onset non-decreasing, inside the primary stream \\
\code{events/r$k$/value, label\_id} & int64, int32 $(E,)$ & raw trigger code or 0; index into \code{vocab} \\
\code{events/r$k$} attrs & & \code{n\_events}, \code{digest} (\code{sha256:} over onset, duration, value, label\_id, vocab, each length-prefixed), \code{vocab} \\
root attr \code{events\_revision} & int & which revision is live; every lower number is an empty tombstone \\
\code{signals/<stream>} attrs & & \code{sfreq}, \code{n\_channels}, \code{n\_samples}, \code{unit}, \code{storage}, \code{compression}, \code{t0\_offset}, \code{highpass}, \code{lowpass}, \code{n\_clipped} (MUST be 0), \code{content\_hash} \\
\bottomrule
\end{tabular}
\end{table}

Conformance is checked by \texttt{validate\_dataset(path, verify\_hashes=True)}. The validator enforces: (D1) root attributes, group structure, and two-level consolidation; (D2) index columns aligned with recording groups; (D3) split and subject indices within declared ranges; (D4) no split-unit leakage at the declared split level; (D5) per-recording conformance, including exactly one live, digest-verified event revision, tombstones for all lower revision numbers, unique channel names, a consolidated snapshot identical to the node document, and consistency with the dataset vocabulary and index row; and (D6, opt-in) content-hash verification.

\section{Canonical channel}\label{app:canonical}

Ids 0--90 define the extended 10--10 scalp space on an ideal sphere of radius 87.5,mm in head coordinates, including \code{Nz}, \code{Fp1/Fpz/Fp2}, AF--PO rows, \code{O1/Oz/O2/O9/O10}, \code{Iz}, earlobes \code{A1/A2}, and mastoids \code{M1/M2}. Midline inclinations are $72^\circ,54^\circ,36^\circ,18^\circ,0^\circ,-18^\circ,-36^\circ,-54^\circ,-72^\circ$ from Fp through O; ring azimuths are $18^\circ$--$162^\circ$ in $18^\circ$ increments. Lateral positions 1/3/5/7 lie at $1/4,1/2,3/4,1$ of the great-circle arc to the ring, with 9/10 positioned 10

Canonicalization additionally defines legacy and vendor aliases: \code{T3/T4/T5/T6}$\to$\code{T7/T8/P7/P8}; AASM \code{LOC/ROC/E1/E2}$\to$ horizontal EOG; TUH \code{EEG <name>-REF/-LE} strips the modality prefix while retaining the reference; and bipolar \code{<a>-<b>} maps to electrode $a$ with reference $b$. Trailing padding and duplicate-name suffixes are removed before lookup. Appending ids is a minor table revision; reordering is major because model input embeddings are indexed by canonical id.

\section{Provenance example}

\begin{footnotesize}
\begin{verbatim}
{"source":   {"file": ".../chb01_03.edf", "format": "EDF", "reader": "edfio",
              "equipment": "X", "startdate": "from the EDF header",
              "timezone": "unknown; EDF has none, stored as UTC",
              "streams": [{"stream": "eeg", "rate_hz": 256.0, "channels": [...],
                           "transducer": [...], "prefiltering": [...],
                           "physical_dimension": [...]}]},
 "pipeline": [{"name": "from_edf", "scaling": "native", "n_streams": 1},
              {"name": "set_bad", "stream": "eeg", "channels": ["T8-P8-1"],
               "note": "duplicate of T8-P8-0", "at": "2026-08-29T01:14:02+00:00"},
              {"name": "append_events", "n_added": 1, "note": "seizure, from the summary",
               "events_revision": 1, "at": "2026-08-29T01:14:03+00:00"}],
 "software": {"eegzarr": "0.3.0", "zarr": "3.3.0", "numpy": "2.2.6"},
 "written":  "2026-08-29T01:14:03+00:00"}
\end{verbatim}
\end{footnotesize}

The converter retains header metadata required for provenance and interpretation, but excludes the EDF patient field because it may contain direct identifiers. Recording start times are treated as quasi-identifiers; converters should shift dates consistently per subject and keep the shift key outside the store. \code{subjects/} contains only run-required fields, while licenses are preserved in \code{sources[]}. Authentication, authorization, and encryption remain responsibilities of the storage layer.

\section{Additional measurements}\label{app:more}

\begin{table}[h]
\caption{Metadata ablation for 200 recordings: metadata queries over rate, length, split, canonical mask, and vocabulary under injected per-request latency; sidecars incur latency per file, Zarr per store request.}
\label{tab:metadata}
\centering\footnotesize
\begin{tabular}{@{}lrrrrr@{}}
\toprule
representation & files & requests & 0\,ms (NVMe) & 1\,ms (NFS-like) & 10\,ms (object store) \\
\midrule
BIDS sidecars (stdlib parsers) & 601 & 601 & 140\,ms & 969\,ms & 6502\,ms \\
Zarr, per-recording channel tables & 4952 & 2614 & 6722\,ms & 9387\,ms & 28252\,ms \\
\textbf{Zarr, dataset index (two-level)} & 4952 & \textbf{14} & \textbf{36\,ms} & \textbf{48\,ms} & \textbf{152\,ms} \\
\bottomrule
\end{tabular}
\end{table}

\begin{table}[h]
\caption{Open / epoch / window timings (median of five warm runs; cells are ZarrFM / EDF / FIF in ms). Sizes use native int16 codes with per-stream byte shuffle (order-0 entropy over stored codes); after shuffle selection the ZarrFM sizes are 20.9, 1.29, and 26.7\,MB. Epochs: 50 of $-0.2\ldots0.6$\,s on a 5\,s grid. Windows: $8{\times}10$\,s, all channels. Sleep-EDF timings omitted (mixed-rate / sparse-channel case).}
\label{tab:real}
\centering\footnotesize
\setlength{\tabcolsep}{3.5pt}
\begin{tabular}{@{}lrrrrrr@{}}
\toprule
recording & EDF & \nz & entropy & open & epochs & windows \\
\midrule
CHB-MIT, 23\,ch, 256\,Hz, 1\,h & 42.4\,MB & \textbf{22.7\,MB} & 8.7\,bit & \textbf{4.7}/24.5/19.6 & \textbf{11}/76/61 & 49/\textbf{15}/\textbf{14} \\
EEGBCI, 64\,ch, 160\,Hz, 2\,min & 2.6\,MB & \textbf{1.4\,MB} & 7.7\,bit & \textbf{4.7}/9.8/6.6 & \textbf{12}/35/16 & 47/\textbf{16}/\textbf{10} \\
Sleep-EDF & 48.3\,MB& \textbf{26.3\,MB} & 9.3\,bit & \textbf{4.5}/45.6/467.4 & \textbf{15.4}/1779.4/1768.2 & \textbf{34.5}/288.2/308.9 \\\bottomrule
\end{tabular}
\end{table}

\begin{figure}[h]
\centering
\includegraphics[width=\textwidth]{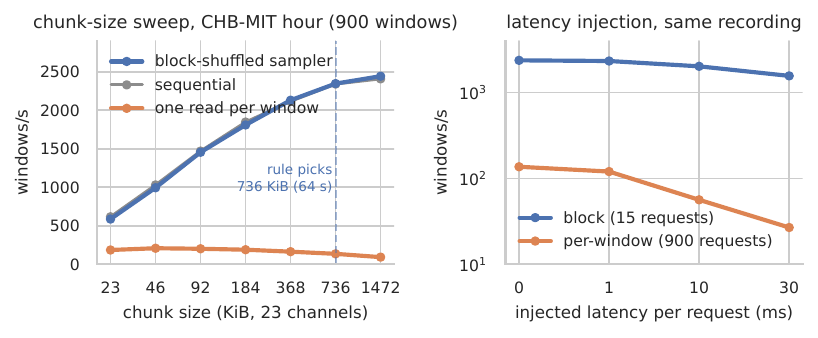}
\caption{Block sampler on CHB-MIT (23,ch, 256,Hz; 900 4,s windows). \textbf{Left:} chunk-size sweep: block-shuffled reads track sequential throughput, while per-window reads decline with larger chunks due to repeated chunk decoding; the selected size is 736,KiB (16,384 samples). \textbf{Right:} simulated per-request latency (bandwidth excluded): 15 block-sampler requests versus 900 per-window requests, reducing latency overhead at 30,ms/request from 34\% to 80\%.}
\label{fig:sampler}
\end{figure}

\begin{table}[h]
\caption{Random-window reads on synthetic data (64\,ch, 10\,min, 256\,Hz; 512 random 4\,s windows; one run), the counterpart of Table~\ref{tab:formats}. Full-range 16-bit codes: the size column is a wash where the real recordings are not, and gzip-HDF5 is $3\times$ slower than on real data. The Parquet row is from a second, more loaded run.}
\centering\small
\begin{tabular}{@{}lrrrr@{}}
\toprule
format & size & windows/s & MB/s out & ms/window \\
\midrule
NPY float32 memmap & 39.3\,MB & 7240 & 1898 & 0.14 \\
HDF5 float32, chunked, uncompressed & 39.3\,MB & 4066 & 1066 & 0.25 \\
\nz, block-merged reads & 18.5\,MB & 3221 & 844 & 0.31 \\
FIF (MNE, \code{preload=False}) & 39.3\,MB & 693 & 182 & 1.44 \\
HDF5 float32, gzip-4 & 36.3\,MB & 489 & 128 & 2.05 \\
EDF (MNE export, \code{preload=False}) & 19.7\,MB & 460 & 121 & 2.17 \\
Parquet, materialised 4\,s windows, zstd & 36.2\,MB & 102 & 27 & 9.78 \\
\nz, one read per window & 18.5\,MB & 151 & 40 & 6.62 \\
\bottomrule
\end{tabular}
\end{table}

\begin{table}[h]
\caption{Chunk-size sweep on synthetic data (4 recordings $\times$ 10 min, 64 ch, 600 windows) and the CHB-MIT hour (900 windows); throughput in windows/s. Reads per epoch are 12/600 for synthetic data and 6/900 for CHB-MIT, comparing block and per-window access at each chunk size.}
\centering\small
\setlength{\tabcolsep}{4pt}
\begin{tabular}{@{}rrrrrrrr@{}}
\toprule
& \multicolumn{3}{c}{synthetic, 64\,ch (bytes at 64\,ch)} & & \multicolumn{3}{c}{CHB-MIT, 23\,ch} \\
chunk & sequential & block & per-window & & sequential & block & per-window \\
\midrule
512 (64\,KiB) & 475 & 490 & 168 & & 617 & 588 & 186 \\
1024 (128\,KiB) & 660 & 730 & 190 & & 1031 & 995 & 209 \\
2048 (256\,KiB) & 941 & 1001 & 183 & & 1470 & 1455 & 202 \\
4096 (512\,KiB) & 1201 & 1294 & 167 & & 1847 & 1810 & 190 \\
8192 (1\,MiB) & 1286 & 1377 & 140 & & 2116 & 2131 & 165 \\
16384 (2\,MiB) & 1227 & 1324 & 108 & & 2341 & 2346 & 136 \\
32768 (4\,MiB) & 1136 & 1202 & 74 & & 2408 & 2443 & 94 \\
\bottomrule
\end{tabular}
\end{table}

\begin{table}[h]
\caption{Left: injected latency on the synthetic corpus (600 windows). Right: PyTorch \code{DataLoader} scaling (batch 32, 4\,s windows, four 10-minute recordings, persistent workers), canonical space (112 slots) vs physical space (64 channels).}
\centering\small
\begin{tabular}{@{}lrrrr@{}}
\toprule
latency & block win/s & req. & random win/s & req. \\
\midrule
0\,ms & 1359 & 48 & 143 & 600 \\
1\,ms & 1340 & 48 & 119 & 600 \\
10\,ms & 1020 & 48 & 58 & 600 \\
30\,ms & 684 & 48 & 27 & 600 \\
\bottomrule
\end{tabular}\hfill
\begin{tabular}{@{}rrrrr@{}}
\toprule
workers & canonical win/s & MB/s & physical win/s & MB/s \\
\midrule
0 & 1053 & 483 & 1446 & 379 \\
1 & 668 & 306 & 978 & 256 \\
2 & 1037 & 476 & 1388 & 364 \\
4 & 1407 & 646 & 1676 & 439 \\
8 & 1742 & 799 & 2396 & 628 \\
\bottomrule
\end{tabular}
\end{table}

\paragraph{GPU path.} Batch $64\times(112,1024)$ float32 on the RTX~6000: pageable copy 1206 windows/s (0.55\ GB/s), pinned non-blocking 1259, pinned with $X\odot\text{mask}$ on the device 1383 (0.63\ GB/s); the H2D copy alone sustains 16566 windows/s (7.6\ GB/s). ``Direct-to-GPU'' is true of the tensors' layout and misleading about where the time goes: decompression and the Python loader are the bottleneck, not the transfer.

\paragraph{Chunking \& Coalescing.} Chunking rails ($2\text{--}64\,\text{s}$, $256\,\text{KiB}\text{--}4\,\text{MiB}$, $\le 1/8\text{th}$ duration) prevent partial chunking artifacts and balance sequential versus windowed access on sparse montages. Window requests within $2\,\text{MiB}$ are coalesced into single reads to amortize fixed request overheads.

\section{Sampler guarantees, edit protocol and BIDS mapping}\label{app:protocol}

\paragraph{Sampler.} Let $\mathcal{W}$ be the window table of a split and $\mathcal{B}$ its partition into storage blocks. For epoch $e$ with seed $s$, a permutation $\pi_{s,e}$ of $\mathcal{B}$ is dealt round-robin to shard $j = \text{rank}\cdot\text{num\_workers}+\text{worker}$ of $n=\text{world\_size}\cdot\text{num\_workers}$: shard $j$ owns $\{\pi_{s,e}(b): b \equiv j \pmod n\}$. Hence (i) $\bigsqcup_j \mathcal{W}_j = \mathcal{W}$ with no duplicates; (ii) no chunk is decompressed by two shards; (iii) the order is a function of $(s,e)$ on every rank. Per-window mode deals windows instead of blocks. Each is unit-tested for two ranks $\times$ three workers.

\paragraph{Edit protocol.} Starting from a base $(k, \text{attrs}, \text{table})$ read once: (1)~claim $\texttt{events/r}_{k+1}\texttt{/zarr.json}$ using set-if-not-exists with a nonce, then read it back; a committed number is never reused, so successful verification shows that no one committed on top of $k$; (2)~write the four columns under the new keys and verify their lengths and digest through a fresh handle; (3)~record the edit in provenance with $k+1$; (4)~re-check the claim, then write the root document once with $\texttt{events\_revision}=k+1$ and a snapshot for that revision only; (5)~empty revision $k$ into a tombstone. Flagging a channel requires one data write and one provenance write and does not touch the root; readers fetch the provenance node directly, so its snapshot may lag. \texttt{repair()} tombstones revisions below the live revision, deletes uncommitted claims above it after re-reading the root, recreates missing tombstones, removes provenance steps that reference an uncommitted revision, and reconsolidates with a guarded write. On a local filesystem, editors and \texttt{repair} also take an advisory lock beside the store. The crash matrix injects a failure after every store operation, and concurrency tests use real processes. Adversarial review found and removed four defects that could lose acknowledged writes under concurrency: deriving the claim number from a second root read; freeing a superseded revision key so a stale editor could reclaim it; performing a read-modify-write of the root after the commit in both channel edits and \texttt{repair}, which could overwrite a concurrent commit; and claiming the ``next free number,'' which allowed two editors from the same base to commit.
\paragraph{Mapping BIDS-EEG to \nz.} Entities \texttt{sub/ses/task} map to index columns and recording IDs; \texttt{\_eeg.json} fields \texttt{SamplingFrequency}, \texttt{EEGReference}, \texttt{PowerLineFrequency}, and filters map to \texttt{sfreq}, \texttt{reference}, \texttt{line\_freq}, \texttt{highpass}, and \texttt{lowpass}; \texttt{\_channels.tsv} fields \texttt{name/type/units/status/reference} map to channel attributes and \texttt{bad}, with \texttt{canonical\_id} derived; \texttt{\_electrodes.tsv}/\texttt{\_coordsystem.json} map to \texttt{position} in the head frame; and \texttt{\_events.tsv} fields \texttt{onset/duration/trial\_type/value} map to revision 0, with \texttt{label\_id}/\texttt{vocab} derived and HED stored in provenance. \texttt{participants.tsv} maps to \texttt{subjects/} as an opt-in, minimal table, while \texttt{dataset\_description.json} and the license map to dataset provenance. The reverse export restores an MNE \texttt{Raw} with names, types, bad channels, montage, filters, and annotations, which MNE-BIDS then writes as BIDS-EEG. Fields not represented by BIDS, including canonical IDs, hashes, revisions, the split manifest, and pipeline provenance, are stored in a \texttt{derivatives/} sidecar. The MNE bridge is implemented; the BIDS mapping is proposed.

\section{Data Availability}

All source code for ZarrFM and related experiments is available on \href{https://github.com/reteena/ZarrFM}{GitHub}.


\end{document}